# A SegFormer-based model for spatial Daylight Factor prediction with comprehensive obstruction encoding and regulatory validation

Stanislava Fedorova[1], Alejandro Pacheco Diéguez[1], Christoph Berkmiller[1], Alberto Floris[2], Libny Pacheco[1], and Marie-Claude Dubois[3]

[1] *Upskiller (BIMTech Innovations AB), Solna, Sweden*

[2] *Arkion AB, Sweden*

[3] *Department of Architecture and Built Environment, Lund University (LTH), Lund, Sweden*

**Abstract.** Daylight compliance assessment in residential buildings currently requires time-consuming raytracing simulations and specialist model preparation, limiting its integration into iterative architectural design. This paper presents a machine learning-based methodology for predicting the spatial distribution of the Daylight Factor (DF) directly within Building Information Modelling (BIM) environments, delivering results in seconds without model export or manual curation. The approach addresses five gaps not simultaneously covered by any existing model: comprehensive external obstruction encoding including balconies, own building height, variable material reflectances, non-orthogonal room geometries, and explicit regulatory validation. Two models are developed and validated: a default-materials model using fixed reflectances for preliminary design checks, and a custom-materials model with variable project-specific reflectances targeting final design verification. Both are trained on approximately 20,000 synthetic Radiance-simulated rooms and validated against a held-out synthetic dataset and 102 rooms from 18 real building models. The default-materials model achieves $\delta_1$ = 83.4% and MAE = 0.11% DF on the synthetic dataset, predicting BBR (Swedish) compliance area within 4.85 pp MAE (bias +2.31 pp) and TEK17 (Norwegian) mean room DF within 0.304% DF MAE (bias +0.287% DF) on real buildings. The custom-materials variant shows 1.7–2.4× higher error across all metrics and requires further development before production use. Both models are deliberately designed to concentrate accuracy in the BBR compliance-critical zone (DF < 1%), achieving a sub-1% DF MAE of 0.105% DF and Compliance IoU of 0.883 on real-world buildings. All cases where the model overestimates compliance are genuinely borderline, within 10.7 pp of the BBR threshold (median 3.2 pp) and 0.36% DF of the TEK17 threshold (median 0.10% DF). The default-materials model is recommended as a replacement for preliminary daylight simulations; the custom-materials variant needs to be further developed before it can be used for final compliance verification.

# 1 Introduction

The importance of daylighting for humans has been demonstrated by decades of research [1, 2, 3], leading to its integration into building regulations and environmental certification systems. In Sweden, the Building Regulations (BBR, BFS 2024:8) require that dwellings achieve a Daylight Factor (DF) of at least 1.0% over a minimum of 50% of assessed floor area of all habitable rooms [4]; in Norway, TEK17 requires a mean room DF of at least 2.0% [5]. Both regulations also accept simplified assessment methods — the AF-method (fönsterglasandelen, referenced in SS 91 42 01) in Sweden, and a 45° obstruction angle rule in Norway — each accepted within a defined scope but with documented accuracy limitations (Section 5.3). These frameworks are under pressure from broader regulation reforms aimed at reducing construction costs and increasing housing supply — making it particularly important to demonstrate that compliance can be verified efficiently, without the time and cost burdens that have historically driven arguments for deregulation.

Current daylight compliance workflows involve three distinct simulation stages. At the urban massing stage, the Vertical Sky Component (VSC) is used to assess whether light incident on windows is likely to be sufficient [6, 7]. While useful early on, VSC does not guarantee DF compliance since it ignores window size, position, glazing transmittance, room dimensions and reflectance [8]. This is followed by preliminary DF simulations with default material assumptions, and finally by detailed simulations with project-specific materials prior to building permit application. Several studies confirm that material reflectance can introduce errors of 10–30% in DF predictions [9, 10], though for typical interior surfaces the practical range is constrained: walls and ceilings are almost always near-white, floor reflectances are stable across projects, and glazing light transmittance (LT) can be treated as a simple linear scalar adjustment.

The main bottleneck is not simulation speed alone, but decision speed: how quickly a design change can be validated and a non-compliance identified. Radiance, the reference engine for regulatory compliance, achieves bias below 15% against physical measurements [11, 12] but requires minutes to hours per simulation. Model preparation — exporting the BIM model, simplifying geometry, assigning materials, and curating the result — can take several hours of specialist work per building. Recent tools accelerate computation (ClimateStudio [13] via progressive path-tracing with AI denoising; Cyclops [14] via GPU acceleration), but neither eliminates the model preparation bottleneck.

Machine learning (ML) offers a third approach: encoding a limited set of architectural parameters to train predictive models that deliver results in real time without model export. The aim of this study is to develop and validate such a model operating directly within BIM environments, and to assess whether its accuracy is sufficient to support reliable compliance checks throughout the design process.

Two predictive models were developed: one with fixed default material reflectances (for preliminary design checks) and one with variable project-specific reflectances (targeting final compliance verification). The main research questions are: (1) To what extent can predictive ML models replicate the accuracy of raytracing simulations for DF assessment across diverse real-world architectural configurations? (2) Are the models sufficiently accurate to support reliable preliminary or final compliance verification?

## 2 State of the Art in ML-based Daylight Prediction

Machine learning for daylight prediction has grown rapidly, from early artificial neural networks (ANNs) and support vector machines (SVMs) to recent convolutional neural networks (CNNs), generative adversarial networks (GANs), and ensemble methods [15, 16, 17]. Despite this progress, five important gaps remain across all reviewed models.

**External obstructions.** Some models exclude surrounding context entirely [18]; others include nearby obstructions but omit balconies and distant buildings [19, 20]. This is particularly problematic for regulatory applications: non-compliance is most likely in lower-floor rooms within dense urban environments, precisely where external obstructions reduce sky visibility most. Pacheco Dieguez et al. [21] made progress by encoding all surrounding obstructions with a window-specific decomposition strategy.

**Own building height.** None of the reviewed models encodes the height of the host building. This is crucial in dense urban contexts where lower floors surrounded by obstructions on multiple sides produce systematically less daylight.

**Variable material reflectances.** Most models use fixed default reflectances [21, 22, 23], which is sufficient for preliminary design checks but not for final official compliance verification, where project-specific material inputs are required. The only exception is Li et al. [24], whose multimodal GAN includes reflectance inputs but does not consider external obstructions or provide regulatory validation.

**Non-orthogonal geometries.** All reviewed models, including [21], are restricted to orthogonal rooms and rectangular windows.

**Regulatory validation.** Few studies explicitly evaluate model performance against compliance thresholds such as BBR or TEK17 [15, 16, 25]. Most validation uses only aggregate metrics (MAE, $R^2$), which do not capture whether a room is correctly classified as compliant or not.

Table 1 summarises these gaps. This study addresses all five simultaneously and validates the resulting model against real-world building models.

**Table 1.** Comparison of ML-based daylight prediction models against key modelling aspects.

| **Model** | **Ext. obstr.** | **Own bldg. ht** | **Var. mat.** | **Non-orth.** | **Tilted/curved** | **Reg. valid.** |
|---|---|---|---|---|---|---|
| Galanos (n.d.) [18] | × | × | × | × | × | × |
| Jia (2021) [19] | Part. | × | × | × | × | × |
| He et al. (2021) [20] | × | × | × | × | × | × |
| Le-Thanh et al. (2022) [23] | × | × | × | × | × | × |
| Alsharif et al. (2023) [22] | × | × | × | × | × | × |
| Li et al. (2024) [24] | × | × | ✓ | × | × | × |
| Pacheco Diéguez et al. (2025) [21] | ✓ | × | × | × | × | Part. |
| **This study** | ✓ | ✓ | ✓ | ✓ | × | ✓ |

## 3 Methodology

The model predicts the spatial distribution of the Daylight Factor across the room floor plane at 0.8 m height. It is trained on a large synthetic dataset with Radiance-simulated DF maps as ground truth (GT), and validated against both a held-out synthetic dataset and real-world building models.

### 3.1 Dataset and ground truth generation

The training dataset consists of 20,687 parametric rooms generated in Grasshopper, covering rectangular and L-shaped geometries (orthogonal and non-orthogonal), room widths and depths of 2–11 m, ceiling heights of 2.5–4.5 m, extensive window and balcony variation (widths 0.8–11 m, depths 0.4–2.2 m), and variable façade, context, and material parameters. Rooms were placed into volumetric urban blocks of Stockholm covering diverse typologies, with façade positions selected to ensure a homogeneous distribution of VSC values from 4% to 45%. Window sizes were distributed around the Swedish DF ≥ 1% compliance threshold [26], focusing model capacity on the compliance-critical range.

Two variants were trained: a **default materials model** with fixed standard reflectances (terrain 0.20, floor 0.30, façade 0.30, walls 0.70, ceiling 0.80, frame 0.50, balcony ceiling 0.60; glazing LT 0.67), and a **custom materials model** with variable project-specific reflectances (terrain 0.05–0.40, floor 0.10–0.60, façade 0.10–0.70, walls 0.30–0.90, ceiling 0.50–0.90, frame 0.10–0.80, balcony ceiling 0.30–0.75; glazing LT fixed at 0.67 in both).

Ground truth DF values were computed with Radiance via Ladybug Tools [27] on a 0.25 × 0.25 m grid at 0.8 m height, using high-quality parameters (ab=6, ad=4096, as=4096). With recommended settings, Radiance achieves bias below 15% against physical measurements [11] — establishing the reference frame against which the ML model is interpreted.

Two validation datasets were used: **Dataset 1** (synthetic): 500 rooms held out before training. **Dataset 2** (real-world): 102 rooms from 18 building models provided by two Scandinavian architecture firms, spanning 18 buildings across five urban typologies (row houses, perimeter blocks, semi-open blocks, slab blocks and tower blocks), deliberately selected for wide variability across geometries, window configurations, obstruction contexts and floor heights.

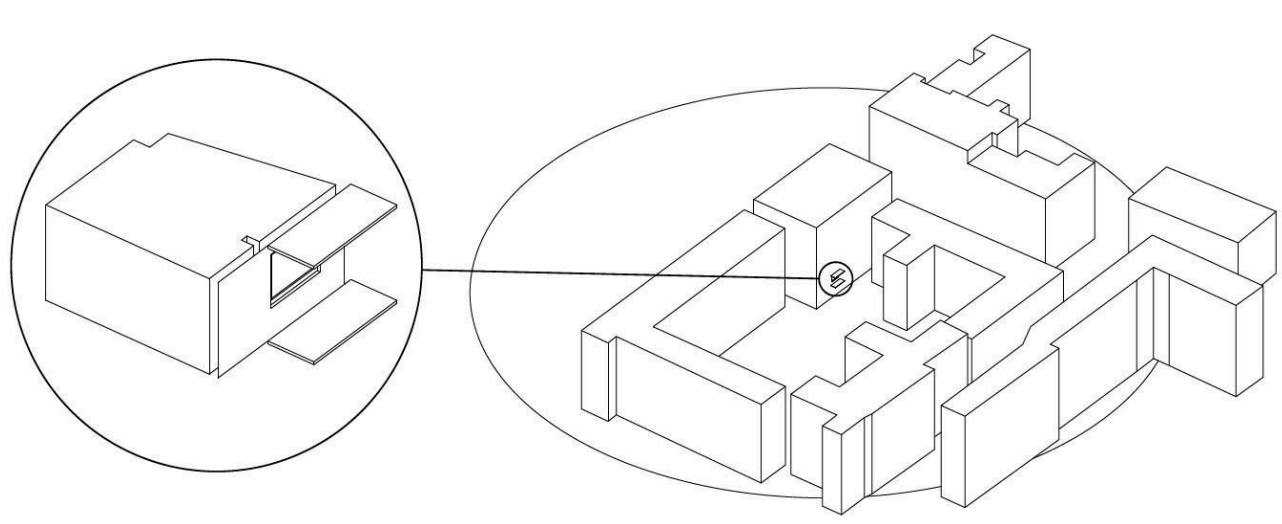

**Fig. 1.** Axonometric view of a training datapoint. Right: the full urban context model with surrounding volumetric blocks. Left (detail): close-up of the target room showing the window, balcony slab, and façade geometry.

### 3.2 Model architecture

The model is based on a modified SegFormer architecture [28], originally developed for semantic segmentation. The encoder uses a Mix Vision Transformer (MiT-B4) with hierarchical feature extraction at four scales (embedding dimensions: 64, 128, 320, 512), combining efficient self-attention with depthwise convolutions. The decoder follows a Global-Local Path (GLP) design, producing a 384 × 384 output map. The final activation is a sigmoid yielding predictions in the 0–9.9% DF range. Training used an L1 (MAE) loss function.

### 3.3 Input encoding

The encoding strategy was designed to extract only the parameters relevant to daylight directly from the BIM model, without requiring model export or specialist preparation. A key architectural decision was to compute each window's daylight contribution independently and sum results in postprocessing — exploiting the superposition principle of daylight. This reduces the learning problem from a multi-window prediction to a series of single-window predictions, and makes it feasible to include complex external obstructions.

The model input consists of two components: a single-channel 128 × 128 geometry mask encoding room shape and window position, and a numerical conditioning vector with scalar parameters injected via learned feature modulation. The conditioning vector includes room dimensions, ceiling height, sill height, window height, frame ratio, floor height above terrain, and own building height. In the custom materials variant, material reflectances are additionally included. External obstructions are encoded as vertical angles from the horizon (surrounding buildings) and zenith (balconies) across 64 planes spanning a 145° field of view (±72.5° from window normal), following [21].

A known limitation: when no sky is visible in a sector, both angles are set to 45° as a sentinel value. This means the model receives identical input for all fully-blocked sectors regardless of the degree of overlap between a balcony above and a building in front — a distinction that can substantially affect daylight levels. Despite this limitation, the real-world validation results show that the model maintains good overall accuracy across obstructed configurations.

### 3.4 Validation metrics

Four metric types were used:

**(a) Threshold accuracy $\delta_1/\delta_2/\delta_3$** — fraction of pixels where max(pred/GT, GT/pred) $< 1.25^n$, adapted from depth estimation [29];

**(b) Quantized Intersection over Union (IoU)** — predicted and GT maps quantized into 5 non-uniform bins (0–0.5%, 0.5–1%, 1–2%, 2–5%, 5–10% DF), concentrating resolution around compliance-critical thresholds, treated as multiclass segmentation;

**(c) MAE** — mean absolute error across all grid points;

**(d) Regulatory compliance** — continuous MAE and bias for the BBR criterion (percentage of floor area with DF ≥1.0%) and TEK17 criterion (mean room DF ≥2.0%), plus Compliance IoU for the sub-1% DF zone.

Sub-1% DF accuracy is reported as a separate metric because the 1% DF threshold is the BBR compliance boundary: a pixel that crosses it in either direction constitutes a direct compliance error. The model predicts each window independently and sums the resulting DF maps in post-processing; in low-DF zones — small or obstructed windows — errors from multiple per-window predictions accumulate at precisely this boundary. Isolating MAE and $\delta_1$ to pixels where GT < 1% DF therefore targets the zone of highest regulatory consequence.

# 4 Results

## 4.1 Synthetic validation (Dataset 1, n=499)

**Table 2.** Validation results on the synthetic dataset (Dataset 1, n=499).

| Metric | df_default | df_custom | Ratio |
|---|---|---|---|
| MAE | 0.113% DF | 0.199% DF | 1.76× |
| RMSE | 0.189% DF | 0.267% DF | 1.41× |
| $\delta_1$ (< 1.25) | 83.4% | 50.8% | 0.61× |
| $\delta_2$ (< $1.25^2$) | 97.1% | 78.2% | 0.81× |
| $\delta_3$ (< $1.25^3$) | 99.3% | 90.5% | 0.91× |
| Quantized IoU | 0.765 | 0.658 | 0.86× |
| BBR compliance MAE | 2.6 pp | 5.6 pp | 2.17× |
| BBR compliance bias | −0.38 pp | −2.47 pp | 6.47× |
| TEK17 compliance MAE | 0.086% DF | 0.173% DF | 2.02× |
| TEK17 compliance bias | −0.005% DF | −0.042% DF | 8.04× |
| Sub-1% DF MAE | 0.051% DF | 0.121% DF | 2.39× |
| Sub-1% $\delta_1$ (±25%) | 83.1% | 42.7% | 0.51× |

The default materials model achieves high accuracy: 83.4% of predicted pixels fall within 25% of ground truth ($\delta_1$) and 99.3% within ~95% ($\delta_3$). BBR compliance is predicted within 2.6 pp (bias −0.38 pp); TEK17 mean room DF within 0.086% DF (bias −0.005% DF) — a negligible underestimation. The custom materials model is consistently worse across all metrics (~1.7–2.2× more error globally). Degradation is most severe in the sub-1% DF zone: $\delta_1$ drops from 83% to 43% and MAE increases 2.4×. Both compliance criteria show larger bias under df_custom: BBR compliance bias increases to −2.47 pp (6.47×) and TEK17 bias to −0.042% DF (8.04×), indicating a systematic underestimation when material reflectances vary from the training defaults. This is the range where reflected light dominates over direct sky component, making variable reflectance most influential. A correlation analysis shows that only walls (r=0.130, p=0.004) and floor (r=0.105, p=0.019) reflectances are statistically significant predictors of error increase — physically consistent with their dominant role in inter-reflections.

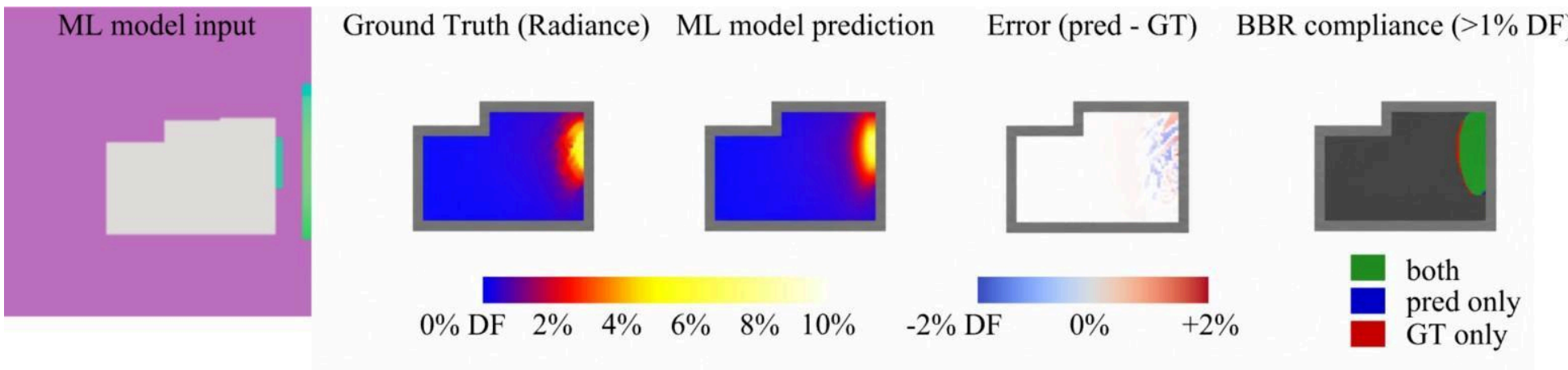

**Fig. 2.** Example prediction pipeline for a synthetic datapoint. From left to right: (a) model input — geometry mask encoding room shape and window position; (b) ground truth DF distribution (Radiance simulation); (c) model prediction; (d) signed error (prediction − GT, ±2% DF scale; blue = underestimation, red = overestimation); (e) BBR compliance overlay (green = compliant in both GT and prediction; red = GT compliant, prediction non-compliant).

## 4.2 Real-world validation (Dataset 2, n=102)

The model performs well on real-world buildings (n=102, 18 building models), evaluated using the default materials model (df_default). All metrics are computed within a 0.5 m wall exclusion zone following EN 17037 and Boverket guidelines. Pixel-level MAE is 0.348% DF (median 0.289%), with 71.5% of predicted pixels within 25% of the ground truth ($\delta_1$) and 98.9% within 100% ($\delta_3$). In the compliance-critical sub-1% DF zone — the range that determines BBR compliance — the model achieves MAE of 0.105% DF and Compliance IoU of 0.883.

**Table 3.** Pixel-level prediction metrics — real-world dataset (n=102, within 0.5 m eval mask).

| Metric | Value |
|---|---|
| MAE | 0.348% DF (median 0.289%) |
| RMSE | 0.519% DF |
| Bias (pred−GT) | +0.287 pp |
| $\delta_1$ (±25%) | 71.5% (median 84.5%) |
| $\delta_2$ (±50%) | 95.1% |
| $\delta_3$ (±100%) | 98.9% |
| Quantized IoU | 0.652 |
| MAE area >1% DF | 4.85 pp |
| Bias area >1% DF | 2.31 pp |

**Table 4.** Sub-1% DF zone metrics — real-world dataset (n=90 rooms with sub-1% DF pixels, within 0.5 m eval mask).

| Metric | Value |
|---|---|
| MAE | 0.105% DF |
| RMSE | 0.120% DF |
| Bias (pred−GT) | +0.062 pp |
| $\delta_1$ (±25%) | 76.4% |
| $\delta_2$ (±50%) | 96.2% |
| $\delta_3$ (±100%) | 99.1% |
| Compliance IoU | 0.883 |

The model estimates the percentage of floor area with DF ≥ 1% (BBR criterion) with a MAE of 4.85 pp and a positive bias of +2.31 pp — a systematic slight overestimation of the compliant area. The mean room DF (TEK17 criterion) is estimated with a MAE of 0.304% DF and a bias of +0.287% DF. Overestimation of compliance occurs only in genuinely borderline rooms — those where the ground truth is already close to the threshold; all cases where the model overestimates compliance are genuinely borderline, with ground-truth compliance rates within 10.7 pp of the 50% BBR threshold (median 3.2 pp) and ground-truth mean DF within 0.36% DF of the 2.0% TEK17 threshold (median 0.10% DF). Figure 3 shows the model performs reasonably well across different configurations, including rooms with deep balconies, buildings in front, compound obstructions, and multiple windows.

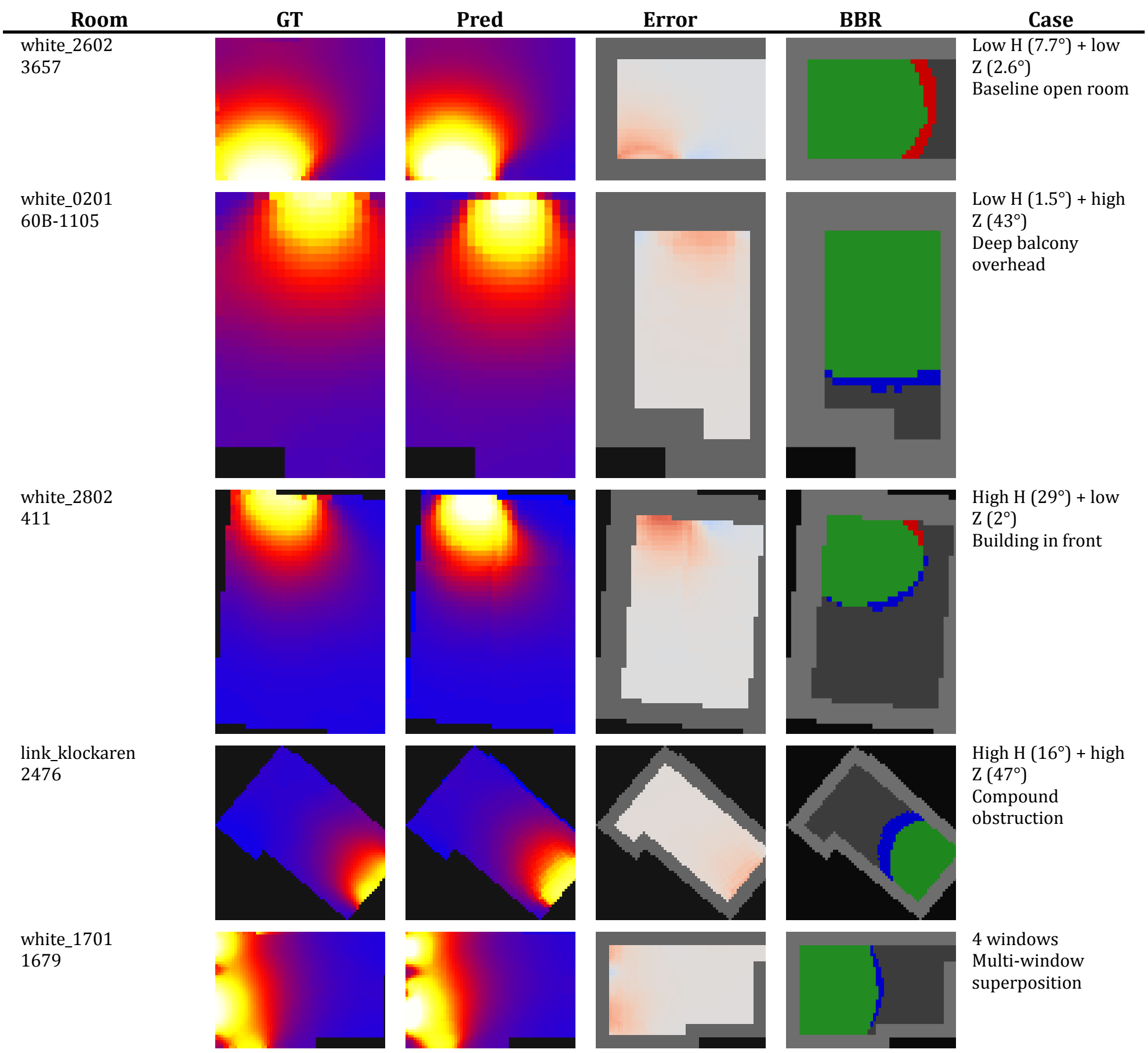

**Fig. 3.** Five representative rooms from the real-world validation dataset (n=102). (A) white_2602/3657 — low horizontal and vertical obstruction (H=7.7°, Z=2.6°), open room. (B) white_0201/60B-1105 — low building-in-front obstruction (H=1.5°), deep balcony overhead (Z=43°). (C) white_2802/411 — building in front, strong horizontal obstruction (H=29°). (D) link_klockaren/2476 — combined horizontal and vertical obstruction (H=16°, Z=47°). (E) white_1701/1679 — four-window room, multi-window superposition. Columns: GT = Radiance ground-truth DF map; Pred = model prediction; Error = pred − GT (RdBu_r ±3 % DF; blue = under, red = over); BBR = compliance overlay (green = both ≥1%; red = GT compliant, pred not; blue = pred compliant, GT not; grey = 0.5 m wall exclusion zone). H = mean obstruction angle from horizon (building in front); Z = mean obstruction angle from zenith (balcony overhead).

# 5 Discussion

## 5.1 Accuracy in context

Two reference frames are relevant for interpreting these results.

The first is Radiance accuracy. Since the model is trained and validated against Radiance simulations, its predictions are bounded by the accuracy of the ground truth. Radiance achieves bias below 15% against physical measurements [11]. On the synthetic dataset, the model's MAE of 0.11% DF is well within this margin. On the real-world dataset (n=102), MAE of 0.348% DF remains within the same order of magnitude — and 0.105% DF specifically in the sub-1% DF zone most consequential for BBR compliance — meaning that, for standard architectural configurations, the ML model can be considered functionally equivalent to a traditional simulation for compliance purposes. For the custom-materials model, MAE increases to 0.199% DF on the synthetic dataset, remaining within the same order of magnitude as Radiance uncertainty, though with degradation concentrated in the sub-1% DF zone (MAE 0.121% DF, $\delta_1$ dropping to 42.7%) — sufficient for exploratory design use but not yet for certification-grade compliance reporting.

The largest errors are concentrated in high-DF zones adjacent to windows — a deliberate consequence of the training strategy, which skewed the dataset distribution towards the compliance-critical range (0–2% DF) and capped ground truth at 10% DF. The regulatory implications of this design choice are discussed in Section 5.2. Furthermore, the Radiance ground truth contains spatial discretisation noise (0.25 m grid); the model's continuous predictions may in some cases be closer to the true underlying distribution than the discretised reference, meaning the reported error metrics slightly overestimate the true prediction error.

The second reference frame is the performance of existing ML models. As shown in Table 1, no previous model has simultaneously addressed external obstructions, own building height, variable material reflectances, non-orthogonal geometries, and regulatory validation. This study integrates all five for the first time, validated against real-world building models.

## 5.2 Regulatory compliance and simplified method comparison

The model is deliberately designed to concentrate accuracy in the sub-1% DF zone, motivated by the superposition-based architecture (§3.3): since each window's contribution is predicted independently and summed in post-processing, only pixels with sub-1% individual contributions have uncertain compliance with the Swedish standard. As a consequence, the positive bias (+2.31 pp for BBR compliance area, +0.287% DF for TEK17 mean) is concentrated in the high-DF zone, leaving BBR compliance predictions largely unaffected — as intended. In practice, overestimation of compliance occurs only in genuinely borderline rooms and by a small margin. The same high-DF overestimation does, however, inflate the predicted mean room DF used by TEK17. This bias (+0.287% DF overall) is not observed on the synthetic held-out set (bias −0.005% DF), because the synthetic dataset was deliberately designed with an over-representation of borderline rooms near the BBR compliance threshold — minimising high-DF examples to concentrate model capacity on the compliance-critical range.

The real-world validation dataset has a more balanced DF distribution, revealing the positive bias in high-DF rooms. Crucially, this overestimation is concentrated in rooms that are already clearly compliant; in genuinely borderline rooms — where compliance assessment matters most — the error is substantially smaller (false positive TEK17 cases have ground-truth mean DF within median 0.10% DF of the 2.0% threshold), making the model reliable for TEK17 compliance evaluation.

The custom-materials model (df_custom) provides meaningful indicative guidance for material sensitivity analysis during design development, but its higher error rates in the sub-1% DF zone ($\delta_1$ dropping to 43%) mean it should not substitute for a final compliance simulation. Where specified materials are equal to or lighter than the default set, df_default — which achieves substantially better accuracy — can be used to demonstrate compliance; where materials are darker, df_custom alone is insufficient for certification and a Radiance simulation is required.

A further argument for the regulatory acceptability of this methodology is the existence of simplified assessment methods already accepted under both Swedish and Norwegian building regulations. In Sweden, the **AF-method** (*fönsterglasandelen*), based on window area to floor area ratio with obstruction correction factors, is referenced in standard SS 91 42 01 and remains an accepted alternative to full Radiance simulations under BBR. In Norway, a preliminary **45-degree obstruction angle rule** is officially accepted under TEK17 to determine whether detailed DF calculations are required. Both methods have documented accuracy limitations: a comparative study of Swedish assessment methods found agreement between GFR and DFP criteria in only approximately 87% of cases, attributed to flaws in their formulations [30], and obstruction angle approaches cannot account for room geometry, window size, or material reflectances. The ML model presented here achieves substantially lower errors — BBR compliance area MAE of 4.85 pp and TEK17 mean room DF MAE of 0.304% DF — while capturing all of these parameters simultaneously. If regulators already accept simplified methods with these known limitations, the accuracy demonstrated here provides a compelling basis for considering this methodology as a legitimate compliance assessment tool.

These findings also have implications beyond individual projects. A recurring argument for deregulation is that compliance is too slow and expensive to enforce efficiently. By demonstrating that compliance can be verified in seconds, directly within the design environment, and with accuracy comparable to traditional simulations for standard configurations, this study removes one of the supporting arguments for weakening a regulation that exists to protect the health and well-being of building occupants.

### 5.3 BIM integration and practical benefits

The methodology addresses the two bottlenecks identified in Section 1.

**Model preparation bottleneck.** The encoding strategy extracts only the parameters relevant to daylight directly from the BIM model, without export, geometry simplification, material assignment, or specialist curation. The window-specific decomposition handles rooms with any number of windows without retraining. The separation of a geometry mask from a scalar conditioning vector makes the system robust to the modelling variability typical of real BIM models in design development.

An incidental benefit of this simplified geometry encoding is its potential to detect errors in Radiance ground truth simulations — a capability with significant value for quality control of final Radiance simulations. The geometry export tools that prepare BIM models for Radiance — whether Grasshopper scripts, dedicated exporters, or plugin pipelines — can introduce two types of artefacts: missing obstructions, where a building element fails to export and is absent from the simulation, and phantom obstructions, where spurious mesh triangulations generate surfaces that do not exist in reality.

Both produce systematic discrepancies between model predictions and Radiance GT that are inconsistent with the room's geometric configuration. When the ML model — which operates on a clean, BIM-native encoding — diverges strongly from Radiance GT in geometrically simple configurations, this is a strong signal of a GT simulation error rather than a model error. During the real-world validation, several such cases were identified and subsequently confirmed as Radiance import failures. Running the ML model alongside the final Radiance simulation as a sanity check therefore provides an independent verification layer that can catch simulation errors before they propagate into compliance reports.

**Computation time bottleneck.** Predictions are delivered in seconds. Room analyses run in parallel on a server, making whole-building analysis feasible at any design stage.

Together, these factors reduce total time associated with daylight compliance by approximately 80–90% compared to traditional workflows, based on preliminary estimations from beta deployments with two Scandinavian architecture firms. Beyond compliance, the tool has a pedagogical dimension — by providing immediate visual feedback as the design evolves, architects gradually develop an intuitive understanding of how window size, room depth, balcony configuration, and external obstructions interact to determine compliance.

This shift in accessibility also raises legitimate questions about the role of the daylight specialist. The specialist's role is not eliminated but transformed: from conducting the full analysis to supervising the process and providing expert judgement in non-standard cases. Furthermore, by lowering the barrier to entry, the tool may expand the overall volume of daylight assessments performed — particularly in countries where daylight analysis is not systematically required — rather than displacing specialist practice.

### 5.4 Limitations and Future Work

The model is restricted to standard architectural configurations: curved windows, tilted walls, skylights, and organic forms cannot be captured by the current encoding and require traditional raytracing. The model currently predicts only the DF — a static metric. Extension to climate-based metrics (Spatial Daylight Autonomy [sDA] and Useful Daylight Illuminance [UDI]) is a priority for future work. A key future direction is replacing or supplementing the synthetic training data with simulations of real building models — and ultimately with physical illuminance measurements — to reduce the domain gap between parametric training conditions and real-world architectural configurations.

Improving df_custom to certification-grade reliability — particularly in the sub-1% DF compliance zone — is identified as a priority for future development.

## 6 Conclusions

This paper presents a machine learning-based methodology for predicting the Daylight Factor spatially within BIM environments, validated against both synthetic (n=499) and real-world (102 rooms from 18 building models) datasets. The main conclusions are:

**Five gaps addressed simultaneously for the first time:** comprehensive external obstruction encoding including balconies, own building height, variable material reflectances, non-orthogonal room geometries, and explicit validation against regulatory compliance thresholds.

**Two workflow bottlenecks eliminated:** (1) BIM-native encoding removes the need for model export, manual curation, and specialist preparation — and as a secondary benefit, enables detection of Radiance simulation errors, providing an independent quality check for final compliance simulations; (2) predictions in seconds with parallel room analysis make whole-building fast assessments feasible at any design stage.

**Accuracy connected to regulatory compliance.** On the real-world dataset (n=102, default materials), pixel-level median error is 0.289 % DF and Compliance IoU is 0.883. All cases where the model overestimates compliance are genuinely borderline — within median 3.2 pp of the BBR threshold and median 0.10 % DF of the TEK17 threshold. This accuracy is within the 15% error of raytracing simulations against physical measurements, and substantially outperforms the AF-method (Sweden) and the 45° obstruction angle rule (Norway) — both already accepted under their respective regulations — while simultaneously capturing room geometry, window size, material reflectances, and external obstructions. The custom-materials variant shows higher error (sub-1% $\delta_1 = 42.7\%$), with degradation in the compliance zone insufficient for certification-grade reporting.

**Model usage recommendations.** The default-materials model (df_default) is recommended as a replacement for preliminary daylight simulations, including VSC checks and early-stage DF assessments, and for final compliance verification where project materials are equal to or lighter than the default set. The custom-materials model (df_custom) is suited for design-stage material sensitivity analysis but should not substitute a final compliance simulation; where project materials are darker than the default set, a Radiance simulation is required for certification. Non-standard configurations — large rooms, double-height spaces, curved or organic forms, and tilted windows or walls — also require Radiance until models covering these conditions are developed.

**Broader implications.** By providing immediate feedback at any design stage, the tool allows likely non-compliance to be detected early — so that when the final simulation is run, any discrepancies are minor and localised rather than requiring fundamental redesign. The results also support the protection of daylight regulation: by demonstrating that compliance can be verified in seconds directly within the design environment, this study removes one of the supporting arguments for deregulation, and actively supports architects in finding solutions that achieve both high density and adequate daylight.

## Contributions and Acknowledgements

Stanislava Fedorova and Alejandro Pacheco Diéguez contributed equally to this work and share first authorship. S. Fedorova led the model architecture design, training pipeline, and ML experimentation. A. Pacheco Diéguez coordinated the project, led the problem framing, dataset generation, obstruction encoding, BIM integration, validation framework and manuscript writing. C. Berkmiller developed the Revit plugin interface. A. Floris contributed to model training. L. Pacheco contributed to project coordination. M.-C. Dubois contributed to the regulatory framework and manuscript review.

The authors gratefully acknowledge the financial support of Bertil och Britt Svenssons Stiftelse för Belysningsteknik (Belysningsstiftelsen), which funded the development of the first predictive model with default materials, the pilot tests with architecture firms, the initial interface enabling practical testing, and a web platform making the tool available to municipalities and smaller architecture practices, as well as part of the writing of this article. Additional support for the further development of the ML model and its commercial deployment was provided by Vinnova (Innovativa Startups), ALMI, and Scaleway. The authors also thank Peter Leuchovius (White Arkitekter) and Arno De Ryst (LINK Arkitektur) for providing real-world building models and their valuable feedback during beta testing.

## AI Disclosure Statement

The authors used Claude (Anthropic, model: claude-sonnet-4-6) as an AI-assisted writing tool to support manuscript drafting and editing. All content was reviewed, verified, and approved by the authors, who take full responsibility for the accuracy and integrity of the published work. No AI tools were used for data analysis, figure generation, or model development.

# Appendix A — Synthetic dataset: parameter distributions

This appendix presents the parameter distributions of the full training dataset (n = 20,687 rooms). Figures A.1–A.4 cover the geometric input parameters used in both model variants. Figure A.5 shows the surface reflectance distributions used exclusively in the custom materials variant. Figure A.6 shows the distribution of the Daylight Factor simulation outputs. Mean (μ) and median values are indicated on each histogram.

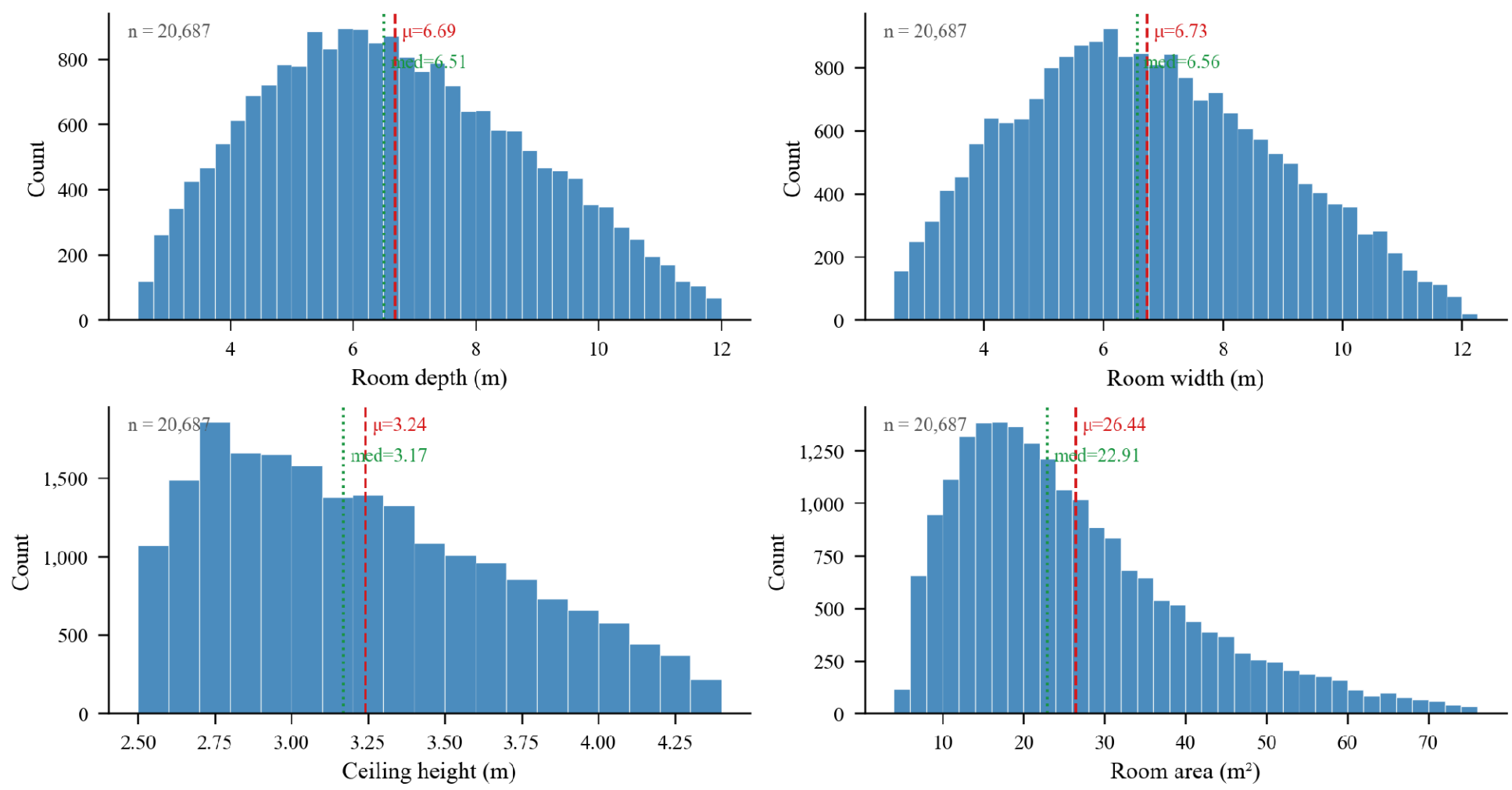

**Fig. 4.** Distribution of room geometry parameters: room depth, room width, ceiling height, and floor area (n = 20,687).

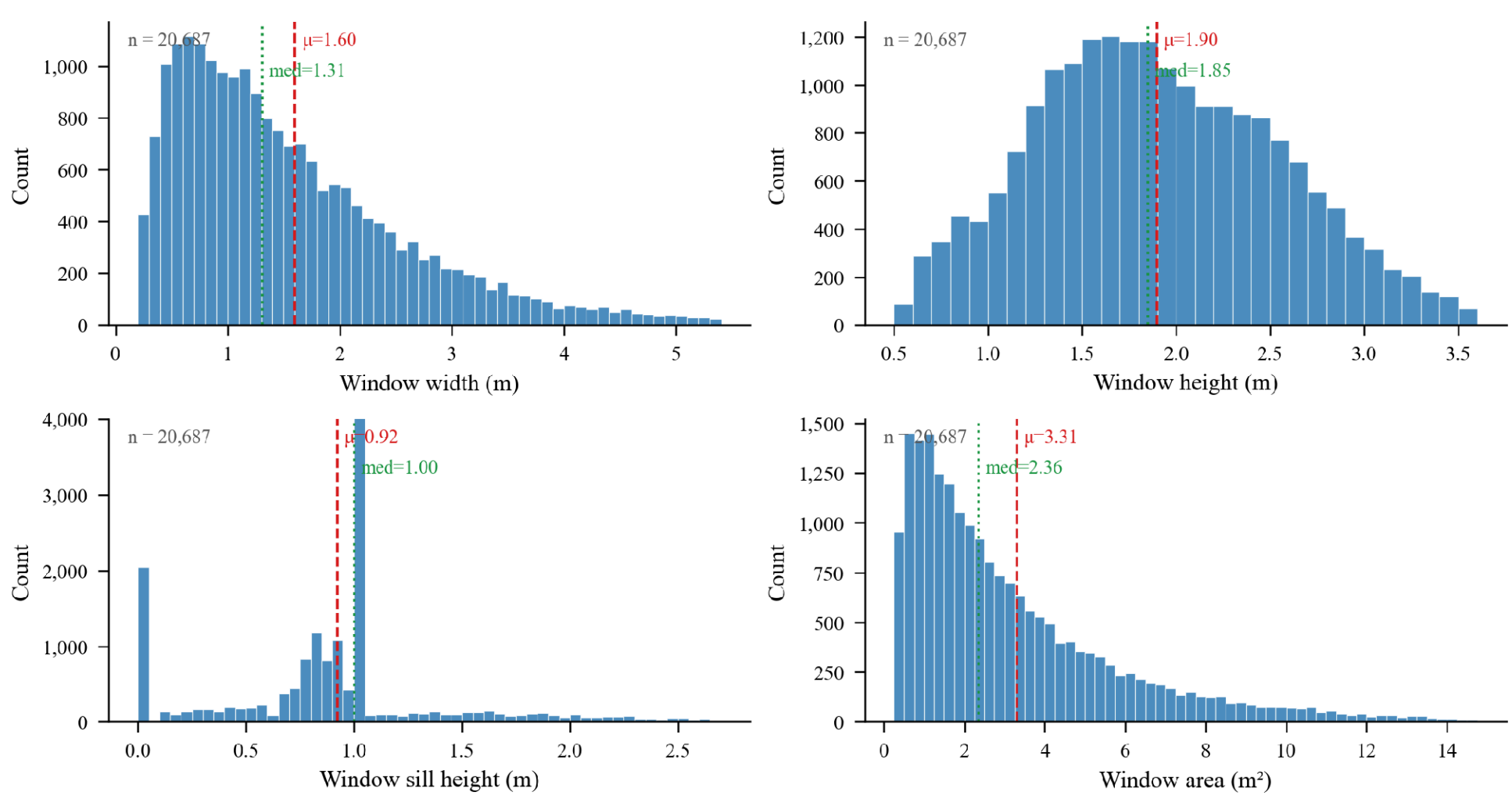

**Fig. 5.** Distribution of window geometry parameters: window width, window height, sill height above floor, and window area (n = 20,687). The bimodal sill height distribution reflects two configurations: floor-to-ceiling windows (sill = 0 m) and standard windows (sill 0.8–1.0 m).

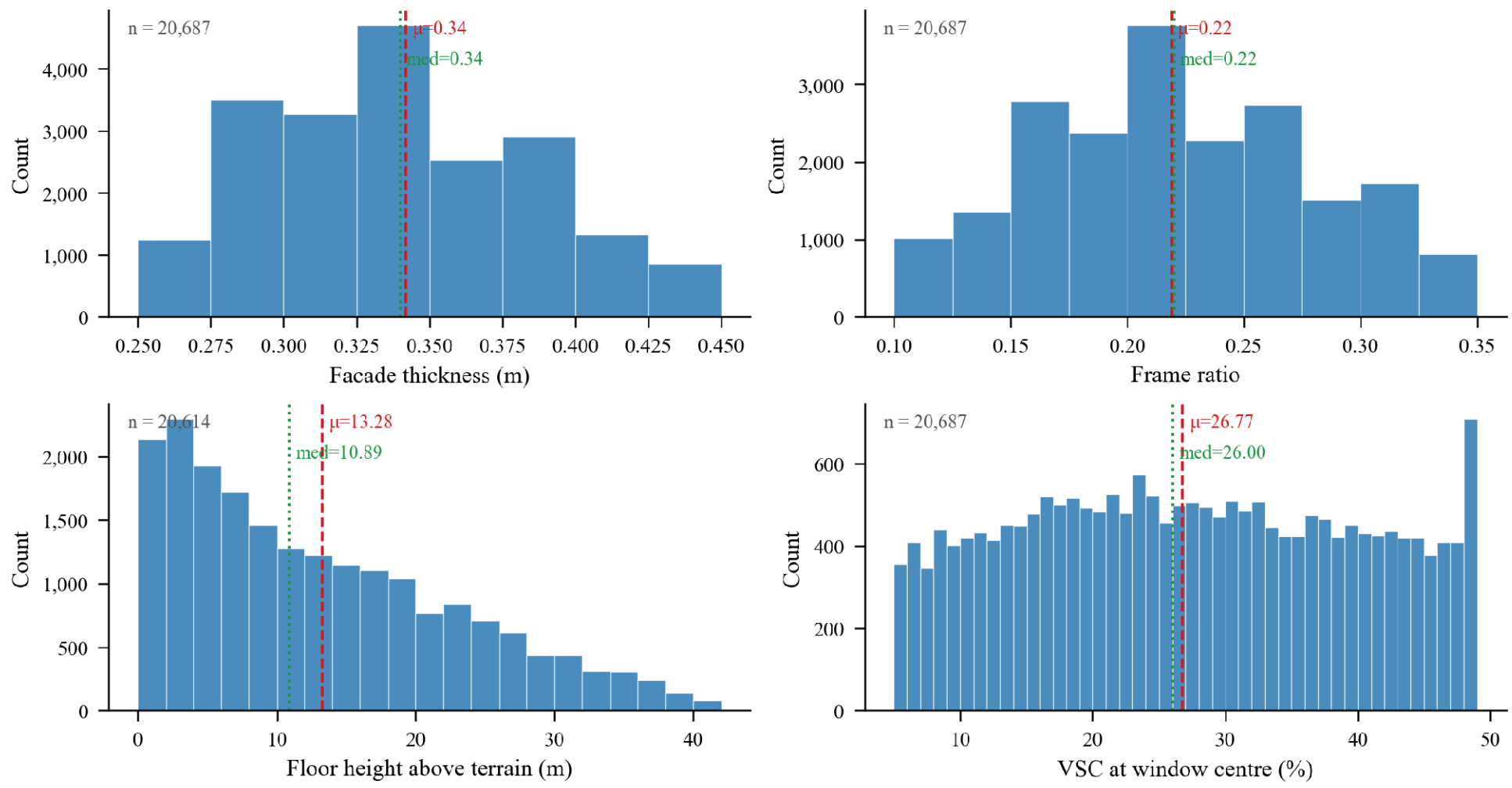

**Fig. 6.** Distribution of envelope and urban context parameters: façade thickness, window frame ratio, floor height above terrain, and Vertical Sky Component (VSC) at window centre (n = 20,687).

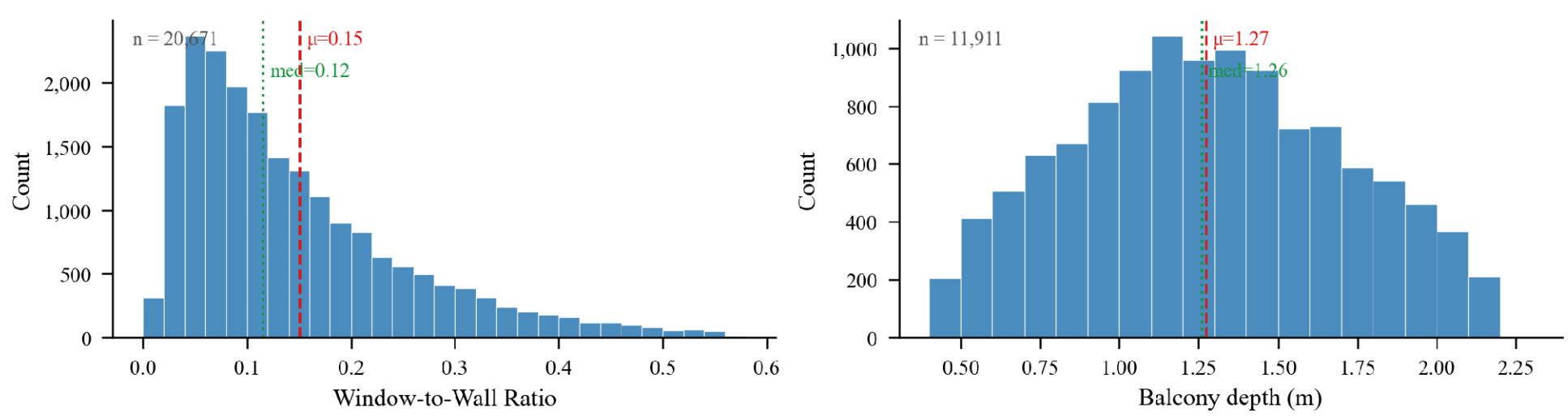

**Fig. 7.** Distribution of Window-to-Wall Ratio (WWR) across all rooms (n = 20,687) and balcony depth for rooms with a balcony (n = 9,674). The WWR distribution reflects single-window configurations; effective total WWR in multi-window rooms is higher.

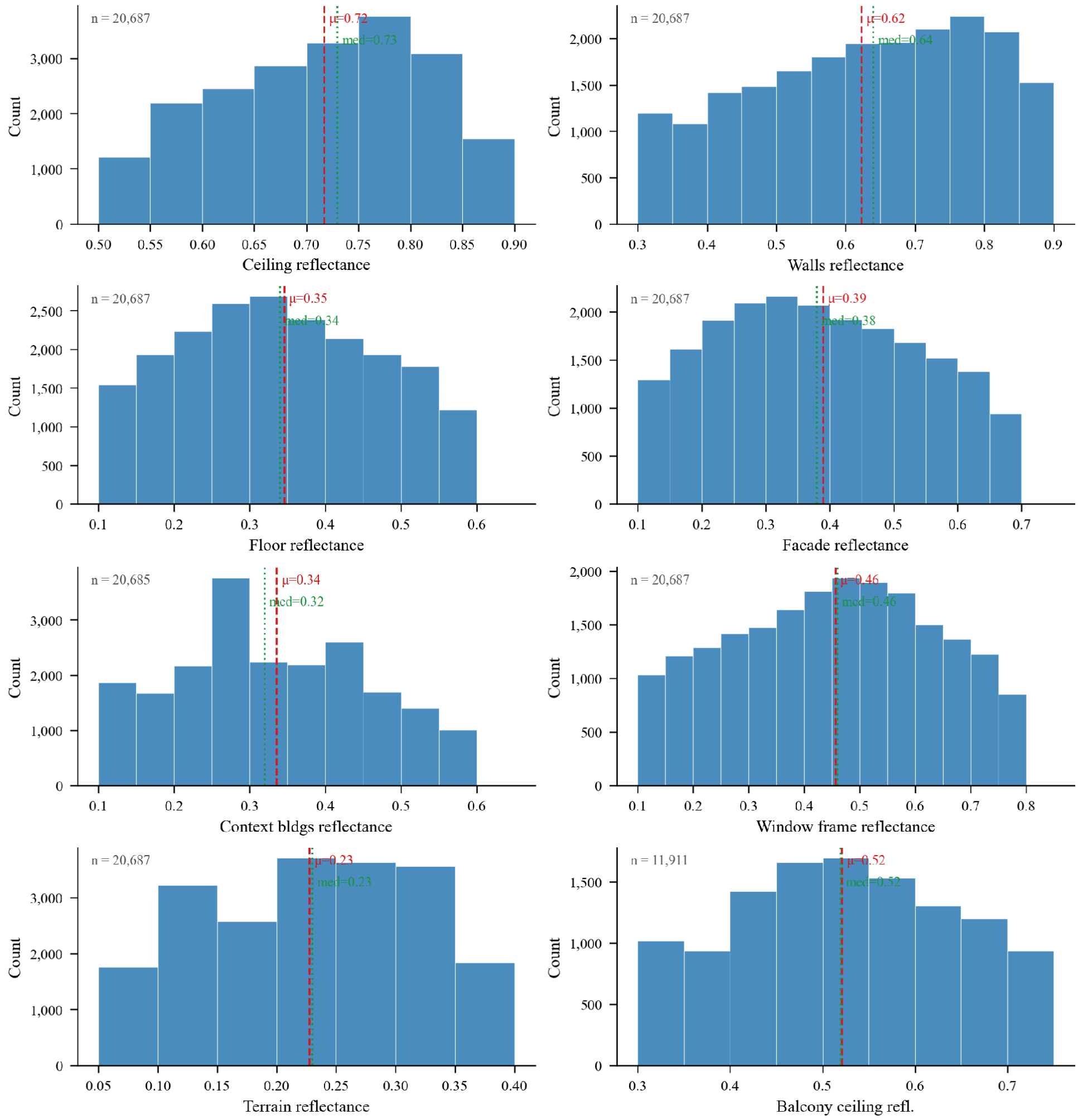

**Fig. 8.** Distribution of surface reflectance values for the custom materials variant: ceiling, walls, floor, façade, surrounding buildings, window frame, terrain, and balcony ceiling. Balcony ceiling shown for rooms with a balcony only (n = 9,674); all other surfaces n = 20,687.

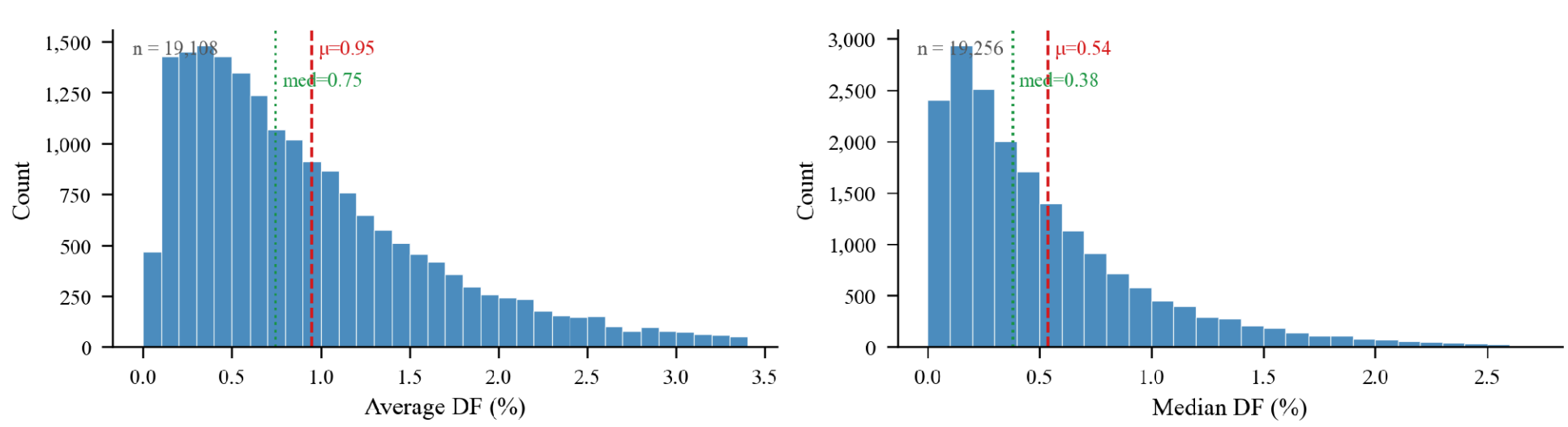

**Fig. 9.** Distribution of Daylight Factor simulation outputs: spatial average DF and spatial median DF per room, computed with Radiance under CIE overcast sky. Values above 3% (< 1% of rooms) excluded for readability.

## Appendix B — Custom materials model: reflectance analysis

This appendix provides supporting detail for the custom materials model (df_custom) results reported in Section 4.1. Table B.1 reports a correlation analysis of which material parameters drive the error increase observed in df_custom relative to df_default across the synthetic test set (n = 499).

**Table 5.** Pearson correlation between surface reflectance and per-sample MAE increase (df_custom − df_default), n = 499.

| Surface | r (MAE) | p (MAE) | r (BBR MAE) | p (BBR MAE) |
|---|---|---|---|---|
| Walls | 0.130 | **0.004** | 0.174 | **<0.001** |
| Floor | 0.105 | **0.019** | 0.122 | **0.007** |
| Ceiling | 0.069 | 0.123 | 0.076 | 0.092 |
| Window frame | 0.033 | 0.459 | 0.074 | 0.098 |
| Façade | 0.048 | 0.287 | −0.006 | 0.890 |
| Terrain | 0.023 | 0.605 | 0.055 | 0.219 |
| Surrounding buildings | 0.040 | 0.370 | 0.046 | 0.305 |

*Source: model_research/eval_outputs_inset/reflectance_analysis/correlations.csv*

## Appendix C — Real-world validation dataset: summary

This appendix summarises the geometric, contextual, and material parameters of the 102 rooms with complete data (n = 18 buildings, after exclusion of 5 anomalous rooms). The 18 buildings contribute between 2 and 13 rooms each (mean 5.7), distributed across five urban typologies: row houses (2 buildings, 6 rooms), perimeter blocks (4 buildings, 25 rooms), semi-open blocks (4 buildings, 29 rooms), slab blocks (5 buildings, 30 rooms), and tower blocks (3 buildings, 12 rooms).

**Table 6.** Surface reflectance values assigned across the 18 real-world validation buildings for the custom materials model variant.

| Surface | Min | Max | Mean | Median |
|---|---|---|---|---|
| Floor / sill / balcony floor | 0.12 | 0.58 | 0.37 | 0.39 |
| Walls / niche | 0.31 | 0.89 | 0.62 | 0.62 |
| Ceilings | 0.52 | 0.89 | 0.69 | 0.68 |
| Balcony ceiling | 0.36 | 0.73 | 0.55 | 0.55 |
| Façade | 0.11 | 0.56 | 0.30 | 0.28 |
| Terrain | 0.10 | 0.37 | 0.27 | 0.29 |
| Surrounding buildings | 0.14 | 0.50 | 0.31 | 0.23 |

**Table 7.** Room-level parameters (n = 102 rooms).

| Parameter | Min | Max | Mean | Median |
|---|---|---|---|---|
| Room area (m²) | 5.2 | 29.6 | 14.8 | 14.0 |
| Number of windows | 1 | 4 | 1.7 | 1 |
| Window-to-floor ratio (WFR) | 0.14 | 0.52 | 0.30 | 0.30 |
| Floor height above ground (m) | 0.0 | 16.6 | 4.3 | 3.2 |

**Table 8.** Window-level parameters (n = 170 windows).

| Parameter | Min | Max | Mean | Median |
|---|---|---|---|---|
| Window area (m²) | 0.97 | 6.29 | 2.4 | 2.3 |
| Frame ratio | 0.14 | 0.52 | 0.30 | 0.30 |
| Window sill height (mm) | 0 | 900 | 473 | 600 |
| Window height (mm) | 1310 | 2310 | 1754 | 1610 |
| Avg obstruction angle — horizon (°) | 0 | 51.0 | 21.3 | 22.2 |
| Avg obstruction angle — zenith (°) | 0 | 62.7 | 24.2 | 18.3 |